\pdfoutput=1

\documentclass[11pt]{article}

\usepackage[final]{emnlp2022}

\usepackage{times}
\usepackage{latexsym}

\usepackage[T1]{fontenc}

\usepackage[utf8]{inputenc}

\usepackage{microtype}

\usepackage{adjustbox}

\title{Revisiting Pre-trained Language Models and their Evaluation for Arabic Natural Language Understanding}
\author{Abbas Ghaddar$^{1,}$\thanks{\hspace{3mm}Equal contribution}\hspace{3mm}Yimeng Wu$^{1,}$\footnotemark[1] \hspace{3mm}Sunyam Bagga$^1$ Ahmad Rashid$^1$ Khalil Bibi$^1$ \\  
\textbf{Mehdi Rezagholizadeh$^1$ Chao Xing$^1$ Yasheng Wang$^1$ Duan Xinyu$^2$}\\
\textbf{Zhefeng Wang$^2$ Baoxing Huai$^2$ Xin Jiang$^1$ Qun Liu$^1$ and Philippe Langlais$^3$}\\
$^1$ Huawei Technologies Co., Ltd.\\
$^2$ Huawei Cloud Computing Technologies Co., Ltd \\
$^3$ RALI/DIRO, Universit\'e de Montr\'eal, Canada\\
\texttt{\{abbas.ghaddar,yimeng.wu,sunyam.bagga\}@huawei.com}\\
\texttt{\{ahmad.rashid,khalil.bibi,mehdi.rezagholizadeh\}@huawei.com}\\
\texttt{\{xingchao.ml,wangyasheng,duanxinyu\}@huawei.com}\\
\texttt{\{wangzhefeng,huaibaoxing,jiang.xin,qun.liu\}@huawei.com} \\
\texttt{felipe@iro.umontreal.ca} \\}

\usepackage{amssymb}
\usepackage{pifont}

\newcommand{\bert}{BERT}
\newcommand{\arabert}{AraBERT}
\newcommand{\asafaya}{Arabic-BERT}
\newcommand{\cmlbert}{CAMeLBERT} 
\newcommand{\arbert}{ARBERT}
\newcommand{\marbert}{MARBERT}
\newcommand{\jaber}{JABER}
\newcommand{\cjaber}{Char-JABER}
\newcommand{\saber}{SABER}

\newcommand{\ats}{AT5S}
\newcommand{\atb}{AT5B}
\newcommand{\artb}{AraT5-base}
\newcommand{\arabb}{AraB2B}

\newcommand{\mightmention}[1]{}
\newcommand{\problem}[1]{\textcolor{red}{$\star$}}
\newcommand{\answer}[1]{\textcolor{blue}{$\#$}}
\newcommand{\todoreview}[1]{\textcolor{green}{$@$}}

\usepackage{subcaption}
\usepackage{rotating,csquotes}
\usepackage{xcolor}
\usepackage{multirow}
\usepackage{framed}
\usepackage{booktabs}
\usepackage{amsmath}
\usepackage{upgreek}
\usepackage{ulem}
\usepackage{amsfonts}

\usepackage[most]{tcolorbox}

\newtcbox{\mybox}[1][]{enhanced, colframe=blue, colback=blue!15, 
	frame style={opacity=0.25}, interior style={opacity=0.25}, 
	nobeforeafter, tcbox raise base, shrink tight, extrude by=1mm, #1}

\begin{document}
\maketitle

\begin{abstract}
There is a growing body of work in recent years to develop pre-trained language models (PLMs) for the Arabic language.
This work concerns addressing two major problems in existing Arabic PLMs which constraint progress of the Arabic NLU and NLG fields.First, existing Arabic PLMs are not well-explored and their pre-trainig can be improved significantly using a more methodical approach. Second, there is a lack of systematic and reproducible evaluation of these models in the literature. In this work, we revisit both the pre-training and evaluation of Arabic PLMs. In terms of pre-training, we explore improving Arabic LMs from three perspectives: quality of the pre-training data, size of the model, and incorporating character-level information. As a result, we release three new Arabic \bert{-style} models ( \jaber{}, \cjaber{}, and \saber{}), and two T5-style models (\ats{} and \atb{}).
In terms of evaluation, we conduct a comprehensive empirical study to systematically evaluate the performance of existing state-of-the-art models on ALUE that is a leaderboard-powered benchmark for Arabic NLU tasks,and on a subset of the ARGEN benchmark for Arabic NLG tasks. We show that our models significantly outperform existing Arabic PLMs and achieve a new state-of-the-art performance on discriminative and generative Arabic NLU and NLG tasks. Our models and source code to reproduce of results will be made available shortly.

\end{abstract}

\section{Introduction}

Pre-trained language models (PLMs) such as BERT~\citep{devlin2018bert}, GPT~\cite{radford2018improving_gpt}, and T5~\cite{raffel2019exploring} have become the default standard architectures for modern natural language understanding (NLU) systems in both academic~\cite{ammus_survey_PLMs,min2021recent}  and industrial~\cite{industry-scale-BERTQA,tunstall2022natural,li2021review} settings. On the evaluation side, the community has widely adopted the leaderboard paradigm\footnote{We use the same definition of leaderboard as \citet{ethayarajh2020utility}.} as a reliable and fair tool to track the progress on various NLP tasks~\cite{MehriDialoGLUE2020, wang2018glue, wang2019superglue}.

Recent years have seen tremendous efforts to develop language-specific PLMs~\cite{le2019flaubert,chan2020germans, canete2020spanish, ulvcar2020finest} and leaderboards~\cite{xu2020clue,xu2021fewclue,shavrina2020russiansuperglue,wilie2020_indonlu} for languages other than English. These language-specific PLMs have proven to be more accurate than multilingual ones in monolingual evaluation settings~\cite{martin2019camembert,wei2019nezha,safaya2020kuisail}. Moreover, creating high-quality human-curated benchmarks is considered to be of utmost importance for reliable evaluation~\cite{deyoung2020eraser,kiela2021dynabench}.

For some high-resource languages like Chinese, the community has been able to be on par with English NLU in terms of developing PLMs~\cite{sun2019ernie,sun2020ernie,sun2021ernie, zeng2021pangu} and evaluating them on publicly available leaderboards~\cite{xu2020clue}. However, we find that for other languages like Arabic the community is unfortunately lagging behind.
Despite the wide availability of Arabic PLMs~\cite{abdul2021arbert, antoun2020arabert, nagoudi2022_arat5, inoue2021interplay} and datasets~\cite{zeroual2019osian, el20161, nagoudi-fake-news-dataset}, there are two major issues that constrain the progress of Arabic NLU field.

First, we observe that the latest techniques for improving pre-training~\cite{brown2020language_gpt3, canine_tokenization_free, efficient_pretraining_objectives} are under-explored in the context of Arabic PLMs. In this work, we investigate three ways to improve on the existing Arabic PLMs: quality of the pre-training data, size of the model, and morphology. We propose \jaber{}, a \bert{-base} model pre-trained on high-quality filtered data, that significantly outperforms the best Arabic PLM baseline by 1.5\% on the ALUE leaderboard~\cite{seelawi2021alue}, a newly proposed benchmark with a leaderboard for sequence classification Arabic tasks.\footnote{The Arabic equivalent of GLUE~\cite{wang2018glue}.} 
We also explore two other variants of our model: (i) \cjaber{} which exploits character-level information and (ii) \saber{} which involves a \bert{-large} model, and report further gains in performance.

Second, there is a lack of systematic and reproducible evaluation. As a matter of fact, most of the existing work on Arabic PLMs does not follow the recommended evaluation protocols~\cite{reproducibilitychecklist,chen2022reproducibility} which include extensive hyper-parameter tuning, performing multiple runs on the development set, and reporting performance on hidden test sets. To address this issue, we systematically compare five popular Arabic \bert{}-like PLMs by carefully assessing their performance on ALUE~\cite{seelawi2021alue}. We find that the performance ranking of models drastically changes when measured on dev sets as compared to the leaderboard test sets, thus calling for caution when comparing models without a leaderboard setting.

Furthermore, we extend our work to T5 encoder-decoder models and Arabic generative tasks. We pre-train two T5 small and base models for Arabic: \ats{} and \atb{}. \atb{} achieves state-of-the-art results on several generative tasks~\cite{naous-empathy-dataset-2020, ladhak-etal-2020-wikilingua} by outperforming the recently proposed \artb{} model \cite{nagoudi2022_arat5} both on automatic and human evaluations.
We further observe that T5-based Arabic PLMs perform worse than the BERT-based models on the ALUE benchmark which is in contrast to the powerful performance of T5-models on English language tasks \cite{raffel2019exploring}.
We conclude with a set of suggestions and directions to explore for pushing forward progress in the Arabic NLU community. 

\begin{table*}[!ht]
    \centering 
    \resizebox{\textwidth}{!}{
    \begin{tabular}{l|ccccc|cc} 
    \toprule
    \bf Model &  \bf \asafaya &  \bf \arabert & \bf \cmlbert & \bf \arbert &  \bf \marbert &  \bf \jaber &  \bf \saber\\  
    \midrule
    \bf \#Params (w/o emb) & 110M (85M) & 135M (85M) & 108M (85M) & 163M (85M) & 163M (85M) & 135M (85M) & 369M (307M)\\ 
    \bf Vocab Size & 32k & 64k & 30k & 100k & 100k & 64k & 64k\\ 
    \bf Tokenizer & WordPiece & WordPiece & WordPiece & WordPiece &  WordPiece &  BBPE &  BBPE \\ 
    \bf Normalization & \ding{56} & \checkmark & \checkmark & \ding{56} & \ding{56} & \checkmark & \checkmark\\ 
    \bf Data Filtering & \ding{56} & \ding{56} & \ding{56} & \ding{56} & \ding{56} & \checkmark & \checkmark\\ 
    \bf Textual Data Size & 95GB & 27GB & 167GB & 61GB & 128GB & 115GB & 115GB\\ 
    
    \bf Duplication Factor & 3 & 10 & 10 & - & - & 3  & 3\\ 
    \bf Training epochs & 27 & 27 & 2 & 42 & 36 & 15 & 5 \\
    \bottomrule
    \end{tabular}
    }
    \caption{Configuration of publicly available Arabic \bert{} models and our models: \jaber{} and \saber{}. \arabert{} and \marbert{} did not provide their data duplication factor. \cjaber{} has the same characteristics as \jaber{}.} %
    \label{tab:model_summary} 

\end{table*}

\section{Related Work}
\label{sec:Related Work}

There have been several efforts to improve on the pre-training paradigm by scaling up the model size~\cite{gshard_largemodel, brown2020language_gpt3} and data size~\cite{liu2019roberta}, exploring new pre-training tasks~\cite{efficient_pretraining_objectives, panda-2021-shuffledtoken_pretraining} and model architectures~\cite{lan2019albert, voita-etal-prune-attention-heads}, and support for long input sequences~\cite{choromanski2021_performer_long, longformer}. In this work, we use the original setting of \bert{}~\cite{devlin2018bert} and T5~\cite{raffel2019exploring} models to pre-train our Arabic encoder-only and encoder-decoder models respectively. The overall goal is to be fairly and directly comparable with other existing Arabic PLMs discussed below. 
 
Table~\ref{tab:model_summary} shows the configuration used by popular publicly available Arabic \bert{} models as well as those of \jaber{} and \saber{}. \arabert{}~\cite{antoun2020arabert} and \asafaya{}~\cite{safaya2020kuisail} were amongst the first to pre-train 12-layer \bert{-base} models specifically for Arabic. \citet{abdul2021arbert} proposed two BERT-based models: \arbert{} which is tailored for Modern Standard Arabic (MSA) NLU tasks and \marbert{} dedicated to tasks that include Arabic dialects (especially tweets). 

\arbert{} and \marbert{} are pre-trained on 61GB and 128GB of MSA and tweets data respectively. \citet{inoue2021interplay} go one step further and pre-train a single \bert{-base} model called \cmlbert{}, on 167GB of MSA, dialect and classic Arabic data. The major difference between \jaber{} and these existing Arabic PLMs is that \jaber{} is pre-trained on a high-quality and strictly filtered dataset (115GB out of 514GB). 

A wide range of methods have been proposed lately to enrich PLMs with character-level information, as it has been shown to be beneficial for morphologically rich languages like Arabic~\cite{kim2016character,gerz2018relation,clark2022canine}. \citet{ma-etal-2020-charbertNLM} proposed Noisy Language Modeling, a new unsupervised pre-training objective to learning character representations. \citet{xrayemb2021} proposed their XRayEmb method that involves adding character-level information to existing PLMs without the need for pretraining them from scratch. CharacterBERT \cite{el-boukkouri-etal-2020-characterbert} uses a character-CNN module to learn representations for entire words by consulting the characters of each token, thus avoiding to recourse to word-pieces~\cite{wu2016google_wordpiece}. Our character-enhanced \bert{-base} model \cjaber{} uses a simple and efficient method to inject character-level representations alongside the sub-tokens representations only at the input layer of \bert{}, with minimal additional parameters and no computational overhead.

Recent efforts have also been made to develop benchmarks for Arabic NLU tasks. \citet{abdul2021arbert} proposed the ARLUE benchmark which is a collection of 42 discriminative classification tasks. \citet{nagoudi2022_arat5} proposed the ARGEN benchmark which consists of 19 datasets for generative tasks. However, both benchmarks have certain limitations which make it challenging to meaningfully evaluate Arabic PLMs. For many tasks, the authors use their own train-dev-test splits which are not made publicly available, as of May 10, 2022. In addition, the access to some datasets is not available free of cost. Furthermore, none of the tasks include privately-held test data which is important to ensure that a benchmark is used fairly \cite{wang2018glue}.
Therefore, we adopt the ALUE benchmark~\cite{seelawi2021alue} for evaluating our models on classification tasks because this benchmark has a public leaderboard and includes privately-held test sets for many tasks. For evaluating our Arabic T5 models, we select a subset of generative tasks from the ARGEN benchmark whose results are freely reproducible (see Section~\ref{sec:Datasets}).

\section{Pre-training}
\label{sec:Pretraining}

\subsection{Data Collection and Processing}

We collect pre-training data from the following sources: common crawl (CC), news (NEWS, EL-KHAIR), and Wikipedia (WIKI).\footnote{See Appendix~\ref{sec:Data collection-Appendix} for data source specific details.} Recent studies have highlighted the importance of cleaning up raw pre-training data for achieving better performance on downstream tasks \cite{raffel2019exploring,brown2020language_gpt3}. We developed a set of heuristics (see Appendix~\ref{sec:Filtering Heuristics-Appendix}) for cleaning our Arabic corpora that is able to filter out gibberish, noisy and duplicated texts.

\begin{table}[!htp]
	\small 
	\begin{center}
		\begin{tabular}{l|l|l}
			\toprule
			
			\bf Source & \bf Original & \bf Clean\\
			
			\midrule
			CC & 475GB & 87GB\hspace{3.2mm}(18\%) \\ 
			NEWS & 21GB & 14GB\hspace{3.2mm}(67\%) \\
			EL-KHAIR & 16GB & 13GB\hspace{3.2mm}(82\%)\\
			WIKI & 1.6GB & 1GB\hspace{5mm}(72\%)\\
			\midrule
			Total & 514GB & 115GB (22\%)\\
			\bottomrule
			
		\end{tabular}
	\end{center}	
	
	\caption{Size of our pre-training corpus before and after applying the data cleaning methods. Parentheses indicate the proportion of the remaining data.}
	\label{tab:pretrain_stat}
\end{table}

Table~\ref{tab:pretrain_stat} shows the size of our pre-training corpora before and after data pre-processing. The final pre-training dataset represents only 22\% of the original corpus and is 115GB in size. Although our approach seemingly filters out a large proportion of the dataset, our corpus size is comparable with other models such as \asafaya{} (95GB) and \marbert{} (128GB). Moreover, as we will discuss in Section~\ref{sec:Experiments}, our models are able to significantly outperform other models that used light pre-processing~\cite{safaya2020kuisail,abdul2021arbert}. We also utilise the Arabic text-normalization procedure of \arabert\footnote{\url{https://github.com/aub-mind/arabert/blob/master/preprocess.py}} which involves removing emojis, tashkeel, tatweel, and HTML markup \cite{antoun2020arabert}.

\begin{table*}[!htp]
    \small 
    \centering
    \begin{tabular}{lrrrlclll}
    \toprule
    \bf Task & \textbf{$|$Train$|$} & \textbf{$|$Dev$|$} & \textbf{$|$Test$|$} & \bf Metric & \textbf{$|$Classes$|$} & \bf Domain & \bf Lang & \bf Seq. Len. \\ 
        
    \midrule
    \multicolumn{9}{c}{\textit{Single-Sentence Classification}}\\
    \midrule
    MDD & 42k & 5k & \bf 5k & F1-macro & 26 & Travel & DIAL & 7$\pm$3.7 \\
    OOLD & 7k & 1k & 1k & F1-macro & 2 & Tweet & DIAL & 21$\pm$13.3 \\
    OHSD & 7k & 1k & 1k & F1-macro & 2 & Tweet & DIAL & 21$\pm$13.3 \\
    FID & 4k & - & \bf 1k & F1-macro & 2 & Tweet & DIAL & 23$\pm$11.7 \\
    
    \midrule
    \multicolumn{9}{c}{\textit{Sentence-Pair Classification}}\\
    \midrule
    MQ2Q & 12k & - & 4k & F1-macro & 2 & Web & MSA & 13$\pm$2.9 \\
    XNLI & 5k & - & \bf 3k & Accuracy & 3 & Misc & MSA & 27$\pm$9.6 \\
    
    \midrule
    \multicolumn{9}{c}{\textit{Multi-label Classification}}\\
    \midrule
    SEC & 2k & 600 & 1k & Jaccard & 11 & Tweet & DIAL & 18$\pm$7.8 \\
    
    \midrule
    \multicolumn{9}{c}{\textit{Regression}}\\
    \midrule
    SVREG & 1k & 138 & 1k & Pearson & 1 & Tweet & DIAL & 18$\pm$7.9 \\
    
    \bottomrule
    \end{tabular}
    \caption{Task descriptions and statistics of the ALUE benchmark. Test sets in bold use labels that are publicly available. The average sequence length and standards deviations are computed based on the word count of the tokenized text of the training set.}
    \label{tab:alue_stat}
\end{table*}

\subsection{Our Models}

We pre-train both \bert{-} and T5-style models. \jaber{} and \saber{} stand for \underline{J}unior (12-layer) and \underline{S}enior (24-layer) \underline{A}rabic \underline{BER}t models respectively. They follow the default configuration of ~\citet{devlin2018bert} \bert{-base} and \bert{-large} respectively. We also enhance \jaber{} with character-level representations at the input layer, which we refer to as the \cjaber{} model.

For \cjaber{}, each word is represented as a sequence of characters, and we use a $m$-layer CNN encoder~\cite{chiu2015named,lee2018character} to obtain a continuous vector of character-level representation for each word. The final input representation is obtained by adding those vectors to the original \bert{} input representations (token, segment, and position). Note that all sub-tokens of the same word share the same character-level representation of that word. 
 
\atb{} and \ats{} use the same encoder-decoder architecture and configuration of T5-base and T5-small \cite{raffel2019exploring} respectively. \atb{} is directly comparable with \artb{}~\cite{nagoudi2022_arat5}, the state-of-the-art model for Arabic generative tasks. The configurations and implementation details of our models are listed in Appendix~\ref{sec:BERT-style Implementation Details},\ref{sec:Encoder-Decoder Models Implementation Details}.

\begin{table*}[!htp]

\centering
\resizebox{\textwidth}{!}{
\begin{tabular}{l|cccccccc|c}
\toprule
\textbf{Model} & \textbf{MQ2Q*}& \textbf{MDD}& \textbf{SVREG}& \textbf{SEC}& \textbf{FID}& \textbf{OOLD}& \textbf{XNLI}& \textbf{OHSD} & \textbf{Avg.}\\
\midrule
\multicolumn{10}{c}{\textit{Baselines}}\\
\midrule

\cmlbert & 68.9$\pm$1.1 & 62.9$\pm$0.1 & 86.7$\pm$0.1 & 45.4$\pm$0.5 & 84.9$\pm$0.6 & 91.3$\pm$0.4 & 55.7$\pm$1.2 & 81.1$\pm$0.7 & 72.1$\pm$0.6 \\ 
\asafaya & 73.3$\pm$0.6 & 61.9$\pm$0.2 & 83.6$\pm$0.8 & 42.4$\pm$0.4 & 83.9$\pm$0.6 &  88.8$\pm$0.5 & 66.0$\pm$0.6 & 79.3$\pm$1.0 & 72.4$\pm$0.6 \\
\arabert & 73.5$\pm$0.5 & 61.1$\pm$0.3 & 82.3$\pm$0.9 & 42.2$\pm$0.6 & 85.2$\pm$0.2 & 89.7$\pm$0.4 & 67.2$\pm$0.4 & 79.9$\pm$1.8 & 72.6$\pm$0.6 \\
\marbert & 69.1$\pm$0.9 & 63.2$\pm$0.3 & \underline{88.0$\pm$0.4} & 47.6$\pm$0.9 & 84.7$\pm$0.4 &  91.8$\pm$0.3 & 63.3$\pm$0.7 & 83.8$\pm$1.4 & 73.9$\pm$0.7\\
\arbert & 74.7$\pm$0.1 & 62.5$\pm$0.2 & 83.5$\pm$0.6 & 43.9$\pm$0.6 & 85.3$\pm$0.3 & 90.5$\pm$0.5 & 70.8$\pm$0.5 & 81.9$\pm$2.0 & 74.1$\pm$0.6 \\
\midrule
 \multicolumn{10}{c}{\textit{Ours}}\\
 \midrule
\jaber & 75.1$\pm$0.3 & 65.7$\pm$0.3 & 87.4$\pm$0.7 & 46.8$\pm$0.8 & 84.8$\pm$0.3 & 92.2$\pm$0.5 & 72.4$\pm$0.7 & 85.0$\pm$1.6 & 76.2$\pm$0.7 \\
\cjaber & \underline{76.8$\pm$0.2} & \underline{67.3$\pm$0.2} & 87.5$\pm$0.3 & \underline{47.8$\pm$0.4} & \underline{85.7$\pm$0.2} & \underline{93.3$\pm$0.1} & \underline{72.7$\pm$0.3} & \underline{86.4$\pm$0.5} & \underline{77.2$\pm$0.3} \\
\saber & \bf 77.7$\pm$0.4 & \bf 67.4$\pm$0.2 & \bf 89.3$\pm$0.3 & \bf 49.0$\pm$0.5 & \bf 86.1$\pm$0.3 & \bf 93.4$\pm$0.4 & \bf 75.9$\pm$0.3 & \bf 88.9$\pm$0.3 & \bf 78.5$\pm$0.3 \\
\bottomrule

\end{tabular}}
\caption{\textsc{Dev} performances and standard deviations over 5 runs on the ALUE benchmark. Bold entries describe the best results among all models, while underlined entries show best results among \bert{-base} models. * indicates that the results are on our own MQ2Q dev set.}
\label{tab:dev_alue}
\end{table*}

\section{Experimental Protocol}
\label{sec:Experiments}

\subsection{Datasets}
\label{sec:Datasets}
We evaluate all models on the newly proposed ALUE benchmark~\cite{seelawi2021alue}, a collection of 8 Arabic NLU tasks. The final score is computed as the unweighted average over those tasks. ALUE is powered by a leaderboard\footnote{\url{https://www.alue.org/leaderboard}} with privately-held test sets and we present a brief summary of the ALUE tasks in Table~\ref{tab:alue_stat} (see Appendix~\ref{sec:ALUE Benchmark} for more detailed description).

Unfortunately, there is no equivalent of the ALUE benchmark (a fixed train/dev/test split, a privately-held test, and a leaderboard) for Arabic generative tasks. Therefore, we evaluate encoder-decoder models on a selected set of generative tasks from the ARGEN benchmark~\cite{nagoudi2022_arat5}: Text Summarization (TS), Question Generation (QG), and Question Answering (QA), where the latter is treated as a sequence-to-sequence generative task. In addition, we experiment on a single-turn dialogue task using the Empathetic Dialogue (EMD) dataset~\cite{naous2021empathetic}. A detailed description of the datasets, splits, and evaluation metrics is available in Appendix~\ref{sec:Generative Tasks Datasets}.

\subsection{Baselines} \label{sec:baselines}
On one hand, we compare our \jaber{}, \cjaber{} and \saber{} models with the popular Arabic PLMs: \asafaya{}~\cite{safaya2020kuisail}, \arabert{}~\cite{antoun2020arabert}, \cmlbert{}~\cite{inoue2021interplay}, \arbert{} and \marbert{} \cite{abdul2021arbert}. On the other hand, we evaluate our \ats{} and \atb{} models against the recently proposed \artb{}~\cite{nagoudi2022_arat5} and \arabb{}~\cite{naous-empathetic-bert2bert} models. The latter is an encoder-decoder model initialized with the weights of \arabert{}. \cmlbert{} and \artb{} refer to \cmlbert{-MIX} and AraT5 models in \cite{inoue2021interplay} and \cite{nagoudi2022_arat5} respectively. These models were pretrained on a mix of MSA and tweets (largest possible corpus) and achieve the best overall performance in their respective papers.

\subsection{Implementation Details}
\label{sec:Fine tune Implementation Details}

In order to ensure a fair comparison amongst existing models, we define a systematic evaluation protocol following the recommendations of \citet{2021MovingTR}. The following four-step approach is applied to every model (including the baselines) for each of the ALUE and generative tasks: \textbf{(1)} We conduct extensive hyperparameter-search experiments (e.g. 60 for \bert{} models) to find the best combination of batch size, learning rate, and dropout rate. \textbf{(2)} We use the best found hyperparameter-setting to perform 5 runs with different random seeds. \textbf{(3)} We report the average and the standard deviation on the development set. \textbf{(4)} We use the best-performing models of the development set experiments for the ALUE leaderboard submissions, as well as for reporting the test-set scores of the encoder-decoder models. The best selected hyperparameters for each model on each task, and other implementation details can be found in Appendix~\ref{sec:Fine tuning Implementation Details}.

\section{Results of \bert{-Style} Models}  
\label{sec:results_bert}

\begin{table*}[!htp]
\small
\centering
\resizebox{\textwidth}{!}{
\begin{tabular}{l|cccccccc|c||c}
\toprule
\textbf{Model} & \textbf{MQ2Q}& \textbf{MDD}& \textbf{SVREG}& \textbf{SEC}& \textbf{FID}& \textbf{OOLD}& \textbf{XNLI}& \textbf{OHSD} & \textbf{Avg.}& \textbf{DIAG}\\
\midrule
 \multicolumn{10}{c}{\textit{ALUE Baselines}}\\
\midrule
m\bert{} & 83.2 & 61.3 & 33.9 & 14.0 & 81.6 & 80.3 & 63.1 & 70.5 & 61.0 & 19.0 \\
\asafaya & 85.7 & 59.7 & 55.1 & 25.1 & 82.2 & 89.5 & 61.0 & 78.7 &  67.1 & 19.6 \\
\midrule
 \multicolumn{10}{c}{\textit{Our Private Submissions of Baselines}}\\
 \midrule
 \arabert{} & 89.2 & 58.9 & 56.3 & 24.5 & 85.5 & 88.9 & 67.4  & 76.8 & 68.4 & 23.5 \\
 \asafaya{} & 89.7 & 59.7 & 58.0 & 26.5 & 84.3 & 89.1 & 67.0 & 80.1 & 69.3 & 19.0 \\
 \cmlbert{} & 89.4 & 61.3 & 69.5 & 30.3 & 85.5 & 90.3 & 56.1 & 80.6 & 70.4 & 11.8 \\
 \arbert{} & 89.3 & 61.2 & 66.8 & 30.3 & 85.4 & 89.5 & 70.7 & 78.2 & 71.4 & 24.3 \\
 \marbert{}  & 83.3 & 61.9 & \underline{75.9} & \underline{36.0} & 85.3 & 92.1 & 64.3 & 78.9 & 72.2 & 12.3 \\

\midrule
 \multicolumn{10}{c}{\textit{Ours}}\\
 \midrule
\jaber & \underline{93.1} & 64.1 & 70.9 & 31.7 & 85.3 & 91.4 & \underline{73.4} & 79.6 & 73.7  & 24.4\\ 
\cjaber & 92.0 & \underline{66.1} & 74.5 & 34.7 & \underline{86.0} & \underline{92.3} & 73.1 & \underline{83.5} & \underline{75.3}  & \bf \underline{26.7}\\
\saber & \bf 93.3 & \bf 66.5 & \bf 79.2 & \bf 38.8 & \bf 86.5 & \bf 93.4 & \bf 76.3 & \bf 84.1 & \bf 77.3  & 26.2\\ 
\bottomrule

\end{tabular}
}
\caption{Leaderboard test results (as of 15/04/2022) of experiments on ALUE tasks and their diagnostic dataset (DIAG). Bold entries describe the best results among all models, while underlined entries show best results among \bert{-base} models.}
\label{tab:test_alue}
\end{table*}

\subsection{ALUE Dev}

The performance of all \bert{}-based models including the baselines on the development set of ALUE tasks is presented in Table~\ref{tab:dev_alue}. We report the average and standard deviation of 5 runs with different random seeds. We observe that the variance in performances of the multiple runs is low and is approximately the same on average for all \bert{-base} models, with the exception of OHSD where all models exhibit higher variance. Interestingly, \cjaber{} and \saber{} report a lower variance across the five runs when compared to the \bert{-base} models.

It can be seen that \asafaya{} and \arabert{} have comparable performances (average score of 72.4\% and 72.6\% respectively). This could be due to the similar size of training data used by both models: \asafaya{} was pre-trained on 95GB of text data that was duplicated 3 times (285GB), while \arabert{} was pre-trained on 27GB duplicated 10 times (270GB). While \cmlbert{} outperforms the other baseline models on certain tasks, it achieves the lowest average score of 72.1. This is due to its poor performance on MQ2Q (68.9) and XNLI (55.7), both of which are sentence-pair classification tasks and involve MSA data.

\arbert{} achieves the highest average score of 74.1\% closely followed by \marbert{} (73.9\%). \marbert{} was pre-trained on a large corpus of Arabic tweets and we observe that it performs well on tasks that involve tweet-data. The opposite holds true for \arbert{}.

Our \jaber{} model  significantly outperforms the best  existing baseline model (\arbert{}) by 2.3\% on the average ALUE score. While \marbert{} performs marginally better on the SVREG and SEC tasks, \jaber{} significantly outperforms \marbert{} on all other tasks, particularly the MSA tasks -- XNLI and MQ2Q -- where it achieves a +9.1\% and +6.0\% of gains respectively.

We see further improvements when the \jaber{} model is enhanced with character representations at the input level. \cjaber{} performs better than \jaber{} on all ALUE tasks resulting in a one point jump in the average ALUE score. Moreover, it can be seen that \cjaber{} outperforms \marbert{} on all tasks (except on SVREG) that involve tweets and dialect data, despite not being pre-trained on tweet corpora. 

Character-level information can be crucial for morphologically rich languages like Arabic, where many low frequent dialect words share the same root and meaning as more frequent MSA words. We integrate this information in an unsupervised manner into both pretraining and fine-tuning stages. We do so without adding any computational overhead and without requiring massive amounts of private tweets data (\marbert{}) which can be difficult to obtain for a large section of the research community.

As expected, our \bert{-large} \saber{} model outperforms all the \bert{-base} models on all ALUE tasks (even \marbert{} on SVREG), achieving a 2.3\% and 1.3\% improvements on ALUE average over \jaber{} and \cjaber{} respectively. In our study, it seems that increasing the model capacity is more important than adding character level information for modelling low frequent dialect words. Nevertheless, combining both techniques may further improve the performances, which we leave for future work. 

\subsection{ALUE Test}

Table~\ref{tab:test_alue} shows the performance of all \bert{-based} models on the ALUE leaderboard. The top two rows correspond to the baselines provided by the ALUE authors and the values are directly taken from the leaderboard. The middle and the bottom sections display the performances of our competitors' baselines and our own models respectively. We keep the baseline results private\footnote{We contacted the owners of the ALUE leaderboard to submit the other baseline models in private mode.} since we are not the owners of these models. Figure~\ref{fig:leaderboard} in Appendix includes a screenshot of the leaderboard from May 2022.

Interestingly, we observe that our private submission of the \asafaya{} model achieves an average ALUE score of 69.3\% which is 2.2 percentage points higher than the one available on the ALUE leaderboard. This can directly be attributed to our extensive fine-tuning protocol (described in Section~\ref{sec:Fine tune Implementation Details}). Specifically, the proper tuning of the hyperparameters for our version of the \asafaya{} model resulted in an overall improvement.  

Surprisingly, we also observe that the relative ranks of the baseline models have changed drastically as compared to the dev set (Table~\ref{tab:dev_alue}). \cmlbert{} had the lowest average ALUE score of 72.1\% on the dev set, but it now outperforms \arabert{} and \asafaya{} on the leaderboard test-set. Similarly, \marbert{} outperforms \arbert{} by 0.8\% on the leaderboard while being 0.3\% behind on the dev set. This happens despite our extensive hyperparameter tuning protocol and the fact that we perform multiple runs. This observation underscores the importance of having separate privately-held test sets to determine the actual state-of-the-art rankings for Arabic PLMs in future.

We observe that our models consistently rank at the top for both ALUE test set and dev set. \jaber{} outperforms all other existing Arabic language models achieving an average score of 73.7\%. \cjaber{} outperforms \jaber{} model with a 1.6\% increase in the average ALUE score. \saber{} expectedly further boosts the average score by 2\% compared to \jaber{}, achieving the new state-of-the-art score of 77.3\% on the ALUE benchmark. 

It is interesting to note that the \cjaber{} model is able to outperform the much larger \saber{} model (by 0.5\%) on ALUE's diagnostic data (DIAG), a dataset which is designed to capture the complex linguistic phenomena of Arabic~\cite{seelawi2021alue}. Moreover, it performs better than \jaber{} on all the ALUE tasks (except MQ2Q and XNLI). Therefore, we argue that augmenting language models with character information is a worthy pursuit for Arabic NLU.

\section{Results of Encoder-Decoder Models} 
\label{sec:results_generative}

Table~\ref{tab:dev_alue_t5} shows the performance of our T5 models (\ats{} and \atb{}) and \artb{}~\cite{nagoudi2022_arat5} on the development split of the ALUE tasks. Expectedly, the smaller variant \ats{} achieves a lower average score. The performance of our \atb{} model is very similar to that of \artb{} with both models slightly outperforming each other on four tasks each. 

\begin{table}[!ht]
    \centering
    \resizebox{\columnwidth}{!}{
    \begin{tabular}{l|cc|c}
    \toprule
    \bf Task name &\bf \ats & \bf \atb & \bf \artb \\
    \midrule
    \bf MQ2Q$\star$ & 73.0$\pm$0.1 & \bf 73.7$\pm$0.1 & 70.5$\pm$2.1\\ 
    \bf OOLD & 88.4$\pm$0.2 & 90.0$\pm$0.4 & \bf 90.5$\pm$0.4 \\ 
    \bf OHSD & 81.0$\pm$1.8 & \bf 81.2$\pm$2.1 & 78.3$\pm$1.4 \\ 
    \bf SVREG & 75.6$\pm$1.6 & 78.1$\pm$2.4 & \bf 80.8$\pm$1.3 \\ 
    \bf SEC & 41.3$\pm$0.5 & 43.8$\pm$0.7 & \bf 44.0$\pm$0.6 \\ 
    \bf FID & 82.1$\pm$0.6 & \bf 83.1$\pm$0.5 & 82.3$\pm$0.4 \\ 
    \bf XNLI & 67.9$\pm$0.3 & 72.2$\pm$0.4 & \bf 72.5$\pm$1.5 \\ 
    \bf MDD & 63.1$\pm$0.3 & \bf 64.7$\pm$0.2 & 63.6$\pm$0.2 \\ 
    \midrule
    \bf Avg & 71.5$\pm$0.7 & \bf 73.3$\pm$0.9 & 73.0$\pm$1.0 \\ 
    \bottomrule
    \end{tabular}
    }
    \caption{Dev Alue scores of Arabic T5-style  models. Results on our own MQ2Q dev set are marked by an $\star$.}
    \label{tab:dev_alue_t5}
\end{table}

Moreover, comparing Table~\ref{tab:dev_alue} and Table~\ref{tab:dev_alue_t5}, we observe that T5-style Arabic PLMs perform worse than the BERT-based models on the same ALUE benchmark. This is in contrast to the powerful performance of T5-models on English language tasks \cite{raffel2019exploring}. This observation requires further investigations, and therefor we did not submit our T5 models to the ALUE leaderboard. 

In order to perform a more meaningful evaluation, we also evaluate the Arabic T5 models on four other tasks: Empathetic Dialogue (EMD), Text Summarization (TS), Question Answering (QA) and Question Generation (QG). We present the performances of all T5-based models on QA in Table~\ref{tab:qa_t5}, on QG and EMD in Table~\ref{tab:qg_emd_t5} and on TS in Table~\ref{tab:ts_t5}. Note that we do not experiment on \arabb{} on TS as BERT model is constrained by a maximum input length of 512.

\begin{table}[!ht]
\centering
\small 
\resizebox{\columnwidth}{!}{
    \begin{tabular}{l|cc|cc}
    \toprule
    & \multicolumn{2}{c|}{\textit{\bf QG}} & \multicolumn{2}{c}{\textit{\bf EMD}} \\
    \bf Model & \bf Dev & \bf Test & \bf Dev & \bf Test\\
    \midrule
    \ats  & 7.8$\pm$0.4 & 15.6 & 2.1$\pm$0.1 & 1.9 \\
    \atb & \bf 8.1$\pm$0.1 & \bf 17.0 & \bf 2.3$\pm$0.1 & \bf 2.0 \\
    \artb &  6.7$\pm$0.1 &13.5 & 2.0$\pm$0.0 &  1.8  \\
    \arabb & 4.7$\pm$0.3 & 11.7 & 2.0$\pm$0.0 & 1.8 \\
    \bottomrule
    \end{tabular}
    }
    \caption{BLEU score of T5-style models on the Question Generation and Empathetic Dialogue tasks.}
    \label{tab:qg_emd_t5}
\end{table}

\begin{table}[!ht]
\centering
\resizebox{\columnwidth}{!}{
    \begin{tabular}{l|cc|cc}
    \toprule
    
    & \multicolumn{2}{c|}{\textit{\bf Dev}} & \multicolumn{2}{c}{\textit{\bf Test}} \\
    \bf Model & \bf EM & \bf F1 & \bf EM & \bf F1\\
    \midrule
    \ats  & 36.8$\pm$0.4 & 57.5$\pm$0.3 & 29.2 & 65.1 \\
    \atb & \bf 40.8$\pm$0.7 & \bf 61.6$\pm$1.1 & \bf 31.6 & \bf 67.2 \\
    \artb & 40.2$\pm$0.4 & 61.4$\pm$0.8 & 31.2 & 65.7 \\
    \arabb & 27.3$\pm$2.5 & 47.9$\pm$1.6 & 22.7 & 54.0\\
    \bottomrule
    \end{tabular}
    }
    \caption{F1-score and Exact Match (EM) of T5-style models on the Question Answering task.}
    \label{tab:qa_t5}
\end{table}

\begin{table}[!ht]
\centering
\small 
\resizebox{\columnwidth}{!}{
    \begin{tabular}{l|ccc}

    \toprule
    & Rouge1 & Rouge2 & RougeL \\
    \midrule
    \multicolumn{4}{c}{\textit{\bf WikiLingua Dev}} \\
    
    \midrule
    \ats  & 24.3$\pm$1.3 & 9.5$\pm$0.6 & 21.6$\pm$1.0 \\
    \atb & \bf 26.1$\pm$2.8 & \bf 10.5$\pm$1.6 & \bf 23.2$\pm$2.5 \\
    \artb & 25.0$\pm$0.2 & 10.0$\pm$0.0 & 22.4$\pm$0.2 \\
    \midrule
    
    \multicolumn{4}{c}{\textit{\bf WikiLingua Test}} \\
    \midrule
    
    \ats & 25.2 & 9.9 & 22.4 \\
    \atb & \bf 27.8 & \bf 11.5 & \bf 24.8 \\
    \artb & 25.1 & 10.2 & 22.5 \\
    \midrule
    \multicolumn{4}{c}{\textit{\bf EASC Test}} \\
    \midrule
    \ats & 11.3 & 2.7 & 10.1 \\
    \atb & \bf 12.6 & \bf 3.5 & \bf 11.3 \\
    \artb & 10.7 & 2.7 & 9.3 \\
    \bottomrule
    \end{tabular}
    }
    \caption{T5-style models' performances on the Text Summarization task.}
    \label{tab:ts_t5}
\end{table}

Our \atb{} model significantly outperforms \artb{} on Question Generation and WikiLingua summarization tasks by 3.5 points and 2.7 points respectively. On the remaining QA and EMD tasks, the performance of the two models is similar with our \atb{} model, performing marginally better. Moreover, we observe in Table~\ref{tab:qg_emd_t5} that even our smaller \ats{} model is able to outperform the bigger \artb{} on QG and EMD tasks while achieving comparable scores on TS and QA tasks. This can be very useful for the community for operating in a low latency setting.

Finally, we observe from Table~\ref{tab:qg_emd_t5} and Table~\ref{tab:qa_t5} that the performance of \arabb{} model is worse than all other T5-based models. We believe that the BERT2BERT approach for Arabic response generation adopted in \cite{naous-empathetic-bert2bert} is not well suited for such generation tasks, and it is preferable to pre-train the model from scratch compared to initializing the encoder-decoder architecture with pre-trained weights.

\begin{table}[!ht]
\centering
\small 
    \begin{tabular}{lll|cc}

    \toprule
    &&& \bf Acceptable & \bf Best \\
    \midrule

   \parbox[t]{2mm}{\multirow{3}{*}{\rotatebox[origin=c]{90}{\textit{\bf QG}}}} 
    &&\atb  & \bf 68\%$\pm$10 & \bf 56\%$\pm$12 \\
    &&\artb & 37\%$\pm$11 & 19\%$\pm$14 \\
    &&\arabb & 40\%$\pm$12 & 25\%$\pm$02 \\
    \midrule

    \midrule
     \parbox[t]{2mm}{\multirow{3}{*}{\rotatebox[origin=c]{90}{\textit{\bf EMD}}}} 
    &&\atb  & \bf 53\%$\pm$08 & \bf 50\%$\pm$07 \\
    &&\artb & 50\%$\pm$12 & 37\%$\pm$10 \\
    &&\arabb & 27\%$\pm$04 & 13\%$\pm$05 \\
    \midrule

    \midrule
     \parbox[t]{2mm}{\multirow{2}{*}{\rotatebox[origin=c]{90}{\textit{\bf TS}}}} 
    &&\atb  & \bf 74\%$\pm$08 & \bf 66\%$\pm$05 \\
    &&\artb & 61\%$\pm$12 & 34\%$\pm$04 \\

    \bottomrule
    \end{tabular}
    \caption{Human evaluation performances on 3 generative tasks.}
    \label{tab:human_eval_t5}
\end{table}

The ideal way to measure performance on language generation tasks is to ask humans to evaluate the models' outputs \cite{nlgsurveyevaluation}. Thus, we evaluate our T5-based and \arabb{} models on the three generation tasks of QG, EMD and TS using human annotators. Each of the three models' outputs on the tasks were evaluated by four, native speakers graduate student, annotators. We perform both absolute and relative comparison of the three models. Specifically, we asked the annotators to label a hundred outputs from each model and each task for two scenarios: (1) Acceptable: each model output is labeled for whether it is acceptable (not strictly perfect) to the annotator or not, and (2) Best: where the annotator must pick exactly one best output out of the three ones. In order to mitigate annotation biases, we randomly shuffle, anonymize and rename the three models' outputs.

The results of our human evaluation study for both Acceptable and Best scenarios are presented in Table~\ref{tab:human_eval_t5}. First, we assert that the reported values are reliable as the standard deviation is low (approximately 10\%) across all tasks. Second, we observe that the scores obtained in the human evaluation study (Table~\ref{tab:human_eval_t5}) are much higher than what the corresponding BLEU and ROUGE scores reported in Table~\ref{tab:qg_emd_t5} would suggest. On EMD, for example, our \atb{} model achieves a score of 53\% for the Acceptable scenario as compared to the previously reported BLEU score of 2.0. This is possible because the dialogue generated by the model could be conveying the same tone and emotion as the reference dialogue which led the annotators to mark it as Acceptable, despite not having a high n-gram overlap with the reference dialogue.

Finally, we can conclude from Table~\ref{tab:human_eval_t5} that our \atb{} model was preferred by the annotators for both scenarios on each of the three tasks. The improvement over \artb{} is considerably large for QG and TS tasks as compared to the empathetic dialogues task. On EMD, we observe that only a fraction of all of the models' responses are considered acceptable by the annotators. However, even in that scenario, the annotators pick our \atb{} model as the best-performing one since it is able to produce the most syntactically correct and coherent responses. One reason for the overall low performance on these tasks is the quality of the datasets available for Arabic NLU: the data is not originally in Arabic and the datasets were created via automatic translation from English datasets. Therefore, in order to make meaningful progress in Arabic NLU, we argue that the community needs to curate high-quality datasets dedicatedly for Arabic.

\section{Conclusion}
\label{sec:Conclusion}

In this work, we revisit the pre-training and evaluation of Arabic PLMs. We introduced five new Arabic language models using both BERT- and T5-style pre-training schemes. Our models outperform all existing Arabic models on the generative tasks as well as on the ALUE benchmark, with \saber{} setting a new state-of-the-art on ALUE. 

In order to accelerate progress in Arabic NLU, we advocate for the creation of more \textit{leaderboard-based} benchmarks with privately-held evaluation sets that covers a wide array of tasks. Moreover, we strongly recommend to evaluate Arabic NLU models following a systematic approach similar to the one we propose, with extensive hyperparameter tuning and multiple runs with different random seeds. 

Having met all the other conditions in the Reproducibility Checklist \cite{reproducibilitychecklist}, we will make the pretrained-weights and source code for our models publicly available shortly. In future, we plan to scale up Arabic PLMs to tens (and hundreds) of billions of parameters in an energy-efficient manner~\cite{glamGoogle,chowdhery2022palm} as well as scale up with high-quality pre-training data ~\cite{hoffmann2022training}.

\section*{Limitations}
While we evaluated our models on a diverse set of classification and generative tasks, there are several NLP tasks that were not accounted for in our study. It would be worthwhile to explore others tasks such as named-entity recognition~\cite{benajiba2007anersys} or coreference resolution~\cite{pradhan2012conll}. Also, there are other Arabic PLMs such as ~\cite{talafha2020multi,lan2020empirical} that were not used in our evaluation study. Those models have been reported to underperform the PLMs we have considered as baselines in our study. Yet, there is a slight chance that including them would change the performance ranking in our evaluation.

As the focus of this study in on overall benchmark performances, we also did not assess the robustness of our models on out-of-domain datasets. Finally, our study lacks a qualitative exploration of the datasets and models' error analyses, which we leave for future work. In particular, we want to check the impressive performance of \cjaber{} on ALUE's diagnostic data.

\section*{Acknowledgments}
We thank Mindspore\footnote{\url{https://www.mindspore.cn/}}, a new deep learning computing framework, for the partial support of this work.

\normalem

\bibliography{custom}
\bibliographystyle{acl_natbib}

\clearpage
\newpage
\appendix

\section{Pretraining Details}
\label{sec:Pretraining Details}

\subsection{Data collection}
\label{sec:Data collection-Appendix}

We collect our pre-training corpus from the following four sources:
\begin{description}

    \item \textbf{Common Crawl (CC)}: We used 10 shards of Common Crawl\footnote{\url{https://commoncrawl.org}} data from March to December 2020. After removing non-Arabic text, this dataset is 444GB in size. Additionally, we use the monthly shard of CC from November 2018 provided by the OSCAR project~\cite{suarez2019asynchronous}. We downloaded  the unshuffled version (31GB) from HuggingFace Datasets~\cite{lhoest2021datasets}.
    
    \item \textbf{NEWS}: We use the links provided in the Open Source International Arabic News Corpus ~\cite{zeroual2019osian} to collect 21GB of textual data from 19 popular Arabic news websites.
    
    \item \textbf{EL-KHAIR}: We use the 1.5 billion words Arabic Corpus \cite{el20161} which is a collection of newspaper articles published by 10 Arabic news sources between 2002-2014.
    
    \item \textbf{WIKI}: We use the Arabic Wikipedia dump\footnote{\url{https://dumps.wikimedia.org/}} from June 2021 and extract the text of articles using WikiExtractor \cite{attardi2012wikiextractor}. 
    
\end{description}

\subsection{Filtering Heuristics}
\label{sec:Filtering Heuristics-Appendix}

\begin{enumerate}
    \item Remove sentences with HTML or Javascript code~\cite{raffel2019exploring}. 
    \item Remove sentences if it has less than 70\% Arabic characters. 
    \item Remove sentences with less than 8 words.
    \item Remove sentences with more than 3 successive punctuation marks (excluding dot).
    \item Remove documents with less than 64 words.
    \item Remove long spans of non-Arabic text (mostly English) inside a sentence. We observe that most of these sentences were gibberish/noisy text and not related to the original content.
    \item Represent each sentence by the concatenation of the first and last 3 words. We only consider words that did not include any digits and were longer than 3 characters. Then, we de-duplicate the corpus by only keeping the first occurrence of sentences with the same key.
    \item Discard a document if more than 30\% of its sentences are removed in our filtering pipeline.
\end{enumerate}

\subsection{\bert{-style} Models}
\label{sec:BERT-style Implementation Details}

For tokenization, we use the byte-level Byte Pair Encoding (BBPE)~\cite{wei2021training} training method which considers the text as a byte sequence. This method improves the learning of the representations of rare words and eliminates the out-of-vocabulary problem. We use a vocabulary size of 64K which is comparable to that of \arabert{}, twice the size of \asafaya{} and \cmlbert{}, and 36\% smaller than \arbert{} and \marbert{}. Our \jaber{} and \saber{} models use the same architecture as that of \bert{-base} and \bert{-large}~\cite{devlin2018bert} respectively. The former is a stack of 12 Transformer-encoder layers (768 hidden units) while the latter consists of 24 Transformer-encoder layers (1024 hidden units). 

For \cjaber{}, we first randomly initialize a character embedding lookup table with a vocab size of 160 characters (induced form the pre-training corpus) and a 768 hidden size. Each word is split into a sequence of characters with a maximum character sequence length of 10. We use two 1-D CNN layers with each layer having a filter size of 348 and a sliding window of size 3. Note that we apply a maxpooling layer with a window size of 5 after the first CNN layer. After the second CNN layer, a linear layer is used to map the final representation to the 768 hidden size. Although this architecture adds an additional 700K parameters to \cjaber{}, this has a negligible computational overhead on \jaber{}. 

Following \citet{devlin2018bert}, we pre-train our models on two unsupervised tasks: Masked Language Modelling (MLM) and Next Sentence Prediction (NSP). Specifically for MLM, we use whole word masking with a probability of 15\%. The original tokens are replaced 80\% of the time with the special \textsc{[MASK]} token, 10\% of the time by a random token, and remains unchanged 10\% of the time. We choose a duplication factor of 3: each input sequence generates 3 random sets of masked tokens.

We pre-train our \jaber{} and \saber{} models on 16 servers for 15 and 5 epochs respectively. Each server constitutes 8 NVIDIA Tesla V100 GPUs with 32GB of memory. The distributed training is achieved through Horovod ~\cite{sergeev2018horovod} with full precision. We use the AdamW ~\cite{loshchilov2017decoupled} optimizer with a learning rate decay setting the initial learning rate to 1e-4 with 10,000 warm-up steps. We train with a maximum sequence length of 128, and set the per-GPU batch size to 64 for \jaber{} and 32 for \saber{}. It takes approximately 16 and 32 hours to conclude one epoch for \jaber{} and \saber{} respectively. Finally, the pre-training setting of our \cjaber{} model is identical to that of \jaber{} with the exception of using a smaller initial learning rate of 5e-5.

\subsection{Encoder-Decoder Models}
\label{sec:Encoder-Decoder Models Implementation Details}

Our text-to-text Transformer models \atb{} and \ats{} use the same encoder-decoder architecture as T5-base and T5-small \cite{raffel2019exploring} respectively. The encoder and decoder components of T5-base have similar configuration as that of BERT-base (12 layers) while T5-small is a smaller model with only 6 layers and 8-headed attention. We use the self-supervised denoising objective \cite{raffel2019exploring} to pre-train our models. Specifically, 15\% of tokens are randomly dropped-out from the input and all consecutive spans of such tokens are replaced by a single sentinel token. The expected output is a sequence of these dropped-out tokens separated by the corresponding sentinel token. We train our T5-style models using the same vocabulary and pre-training corpus as that of our \bert{-style} models.

The models are pre-trained on 64 GPU cluster for 200k steps. The pre-training code is based on the PyTorch~\cite{paszke2019pytorch} version of the Transformers library~\cite{wolf2020transformers}. The distributed training is achieved by PyTorch's native distributed training capabilities. We use the Adafactor optimizer~\cite{adafactor2018} with an initial learning rate of 1 and inverse square-root decay until the end of pre-training. 

For both \ats{} and \atb{}, the maximum sequence length is set to 512 for the encoder and 114 for the decoder. We use a per-GPU batch size of 56 and 16 for \ats{} and \atb{} respectively (the maximum batch size that can fit on a single GPU). It important to note that most of our implementation choices (learning rate, optimizer etc.) strictly follow that of \citet{raffel2019exploring} and \citet{nagoudi2022_arat5}. We only differ from AraT5 through the use of a different pre-training corpus and vocabulary.

\section{Fine-tuning Details}
\label{sec:Fine-tuning Details}

\subsection{ALUE Benchmark}
\label{sec:ALUE Benchmark}

ALUE~\cite{seelawi2021alue} consists of a diverse collection of eight Arabic NLU tasks: four single-sentence, two sentence-pair, and one multi-label classification task, as well as one regression task. Five of the eight ALUE tasks are sourced from Twitter data whereas six tasks involve dialectal Arabic. We refer the readers to \citet{seelawi2021alue} for a detailed description of each individual task. 

We note three potential limitations with the ALUE benchmark: (1) the size of training data and average sequence lengths across tasks are smaller when compared with GLUE~\cite{wang2018glue}, (2) the test set labels are public for three tasks: FID, XNLI, and MDD, and (3) development set is not available for three tasks: FID, XNLI and MQ2Q.

Following \citet{seelawi2021alue}, we use the available test set as the development set for FID and XNLI. In order to create the development set for MQ2Q, we use a simple approach: (i) we convert the development set of another task called QQP\footnote{\url{https://www.quora.com/q/quoradata/First-Quora-Dataset-Release-Question}} from English to Arabic using an online translation service, (ii) we pick a random sample of 2k positive and 2k negative instances. We only pick sentence pairs that do not contain any English letters to create a high-quality development set.

\subsection{Generative Tasks Datasets}
\label{sec:Generative Tasks Datasets}

While there is no equivalent for ALUE for generative tasks, \citet{nagoudi2022_arat5} recently introduced the ARGEN benchmark for Arabic natural language generation composed of 7 tasks and 19 datasets. Besides the lack of a public leaderboard and private test sets, there are certain issues with this benchmark. Some datasets are not available publicly (e.g. ARGEN$_{NTG}$), and in some cases, the exact data split is not made public (e.g. ARGEN$_{TS}$). Therefore, we only consider three ARGEN tasks in our evaluation: Question Generation (QG), and Question Answering (QA), Text Summarization (TS). We did not include any tasks that involved non-Arabic text (e.g. translation) since we restrict the scope of this work to a monolingual setting.

Furthermore, we also evaluate our models on the EMpathetic Dialogues (EMD) dataset \cite{naous-empathy-dataset-2020}, which is an Arabic conversational dataset of empathetic conversations curated by translating its English counterpart~\cite{rashkin2019towards}. Table~\ref{tab:generative_datasets} shows the number of instances in the train/dev/test splits for each dataset. The data collection process and evaluation metrics are adopted from ~\cite{nagoudi2022_arat5,naous-empathetic-bert2bert}. Specifically, we use ROUGE~\cite{lin2004rouge} to evaluate models on the TS task and BLEU~\cite{papineni2002bleu} for QA, QG and EMD tasks.

\begin{table}[!ht]
\centering
    \begin{tabular}{c|ccc}
    \toprule
    \bf Task & \bf |Train & \bf |Dev| & \bf |Test| \\
    \midrule
    TS & 23.4k & 2.9k & 2.9k/153 \\
    EMD & 19.5k & 2.8K & 2.5k \\ 
    QG/QA & 101.6k & 517 & 11.6k \\
    \bottomrule
    \end{tabular}
    \caption{Train/Dev/test sizes of the datasets used to evaluate encoder-decoder models. Note that the test set for TS consists of 2.9K articles from WikiLingua \cite{ladhak-etal-2020-wikilingua} and 153 articles from Essex Arabic Summaries Corpus (EASC) \cite{arabicsummaries2010}.}
    \label{tab:generative_datasets}
\end{table}

We adopt the generative format for QA where the input is a pair of passage and question text, and the model is expected to generate the answer. Following \citet{nagoudi2022_arat5}, we use the same dataset for QG as QA except that the input now is a pair of passage and answer text, and the model must generate the corresponding question.
Note that for the summarization task, \citet{nagoudi2022_arat5} did not publish the exact split they used on WikiLingua \cite{ladhak-etal-2020-wikilingua}. We create our own splits by first randomly shuffling the dataset (with seed = 42) and then splitting with the same proportions of 80\% train, 10\% dev and 10\% test. We will make the code publicly available to reproduce our splits and empirical results.

It is important to mention that the performance scores obtained from our re-implementations on TS and EMD tasks are significantly lower than the original scores reported in~\cite{nagoudi2022_arat5} and \cite{naous-empathetic-bert2bert}. This is due to errors in the original implementations. For TS, we found a major error in the calculation of the ROUGE score as the ROUGE tool used by the authors was incompatible with Arabic. For EMD, we found the original BLEU scores to be inflated as the authors compute it on segmented text and not at the word-level (after de-segmentation).

\subsection{Implementation Details}
\label{sec:Fine tuning Implementation Details}
For \bert{-style} models,  we use the AdamW~\cite{loshchilov2017decoupled} optimizer with a learning rate decay. Fixing the number of epochs to 30, we perform grid search with multiple runs to find the best hyperparameters: learning rate from \{7e-6, 2e-5, 5e-5\}, batch-size from \{8, 16, 32, 64, 128\}, hidden dropout from \{0.1, 0.2, 0.3, 0.4\}. For encoder-decoder models, we use the Adafactor~\cite{adafactor2018} with inverse square root decay and pick a learning rate from \{1e-3, 1e-4, 1e-5\}. The fine-tuning code is based on the PyTorch~\cite{paszke2019pytorch} version of the Transformers library~\cite{wolf2020transformers}. We run all experiments on a single NVIDIA Tesla V100 GPU. The best hyperparameters for the generative tasks and ALUE tasks can be found in Table~\ref{tab:hp_gen} and Table~\ref{tab:hp_cross} respectively.

\begin{table}[!th]
    \centering
    \resizebox{\columnwidth}{!}{
    \begin{tabular}{l|c c c c }
   
    \toprule
    \textbf{Model} & \textbf{QA}& \textbf{QG}& \textbf{EMD}& \textbf{TS}\\
    
    \midrule
    \multicolumn{5}{c}{\textit{\arabb}}\\
    \midrule
    batch size & 16 & 16 & 32 & - \\
    hidden dropout & 0.1 & 0.2 & 0.1 & - \\
    learning rate & 1e-03 & 1e-03 & 1e-03 & - \\
    
    \midrule
    \multicolumn{5}{c}{\textit{\artb}}\\
    \midrule
    batch size & 32 & 8 & 8 & 4 \\
    hidden dropout & 0.2 & 0.2 & 0.1 & 0.1 \\
    learning rate & 1e-03 & 1e-03 & 1e-03 & 1e-03 \\
    
    \midrule
	\multicolumn{5}{c}{\textit{\ats}}\\
    \midrule
    batch size & 32 & 16 & 32 & 4 \\
    hidden dropout & 0.1 & 0.2 & 0.1 & 0.1  \\
    learning rate & 1e-03 & 1e-03 & 1e-03 & 1e-03 \\

    \midrule
    \multicolumn{5}{c}{\textit{\atb}}\\
    \midrule
    batch size & 16 & 32 & 16 & 4 \\
    hidden dropout & 0.1 & 0.1 & 0.1 & 0.2 \\
    learning rate & 1e-03 & 1e-03 & 1e-03 & 1e-03 \\

    \bottomrule
    \end{tabular}
    }
    \caption{Best hyperparameters for Arabic encoder-decoder models on the generative tasks.}
    \label{tab:hp_gen}
\end{table}

\begin{table*}[!th]
    \centering
\begin{adjustbox}{totalheight=\textheight-2\baselineskip}
    \begin{tabular}{l|c c c c c c c c }
   
    \toprule
    \textbf{Model} & \textbf{MQ2Q}& \textbf{MDD}& \textbf{SVREG}& \textbf{SEC}& \textbf{FID}& \textbf{OOLD}& \textbf{XNLI}& \textbf{OHSD}\\

    \midrule
	\multicolumn{9}{c}{\textit{\asafaya}}\\
    \midrule
    batch size & 64 & 16 & 16 & 16 & 32 & 32 & 64 & 16\\
    hidden dropout  & 0.1 & 0.1 & 0.1 & 0.1 & 0.1 & 0.1 & 0.1 & 0.1\\
    learning rate  & 2e-05 & 2e-05 & 2e-05 & 2e-05 & 2e-05 & 2e-05 & 2e-05 & 2e-05\\
    
    \midrule
	\multicolumn{9}{c}{\textit{\arabert}}\\
    \midrule
    batch size & 128 & 32 & 8 & 8 & 8 & 32 & 32 & 16 \\
    hidden dropout  & 0.1 & 0.1 & 0.2 & 0.1 & 0.1 & 0.1 & 0.3 & 0.1\\
    learning rate  & 2e-05 & 2e-05 & 2e-05 & 2e-05 & 2e-05 & 2e-05 & 2e-05 & 2e-05\\
    
    \midrule
	\multicolumn{9}{c}{\textit{\cmlbert}}\\
    \midrule
    batch size & 16 & 8 & 8 & 32 & 8 & 128 & 32 & 8 \\
    hidden dropout  & 0.2 & 0.2 & 0.2 & 0.1 & 0.2 & 0.1 & 0.1 & 0.1\\
    learning rate  & 5e-05 & 2e-05 & 2e-05 & 5e-05 & 2e-05 & 2e-05 & 2e-05 & 2e-05\\

    \midrule
	\multicolumn{9}{c}{\textit{\arbert}}\\
    \midrule
    batch size & 64 & 16 & 32 & 8 & 32 & 128 & 32 & 32\\
    hidden dropout & 0.1 & 0.1 & 0.3 & 0.3 & 0.1 & 0.1 &  0.1 & 0.3\\
    learning rate & 2e-05 & 2e-05 & 2e-05 & 2e-05 & 2e-05 & 2e-05 & 2e-05 & 7e-06\\
    
    \midrule
	\multicolumn{9}{c}{\textit{\marbert}}\\
    \midrule
    batch size & 64 & 64 & 16 & 8 & 64 & 64 & 64 & 64 \\
    hidden dropout  & 0.3 & 0.2 & 0.1 & 0.3 & 0.1 & 0.2 & 0.2 & 0.1\\
    learning rate  & 2e-05 & 2e-05 & 2e-05 & 2e-05 & 2e-05 & 2e-05 & 2e-05 & 2e-05\\
    
    \midrule
	\multicolumn{9}{c}{\textit{\jaber}}\\
    \midrule
    batch size & 64 & 32 &  8 & 16 & 32 & 128 & 16 & 32\\
    hidden dropout & 0.3  & 0.2 & 0.1 & 0.1 & 0.1 & 0.2 & 0.1 & 0.3 \\
    learning rate & 2e-05 & 2e-05 & 2e-05 & 2e-05 & 2e-05 & 2e-05 & 2e-05 & 7e-06\\

    \midrule
	\multicolumn{9}{c}{\textit{\cjaber}}\\
    \midrule
    batch size & 64 & 32 &  32 & 16 & 8 & 32 & 64 & 16\\
    hidden dropout & 0.1  & 0.2 & 0.1 & 0.2 & 0.2 & 0.2 & 0.2 & 0.1 \\
    learning rate & 7e-06 & 2e-05 & 2e-05 & 2e-05 & 2e-05 & 7e-06 & 2e-05 & 7e-06\\
    
    \midrule
	\multicolumn{9}{c}{\textit{\saber}}\\
    \midrule
    batch size & 32 & 32 &  8 & 8 & 32 & 32 & 32 & 32\\
    hidden dropout & 0.1  & 0.1 & 0.2 & 0.2 & 0.3 & 0.2 & 0.2 & 0.1  \\
    learning rate & 7e-06 & 2e-05 & 7e-06 & 2e-05 & 2e-05 & 7e-06 & 7e-06 & 7e-06\\
    
    \midrule
	\multicolumn{9}{c}{\textit{\ats}}\\
    \midrule
    batch size & 16 & 32 & 8 & 16 & 16 & 16 & 8 & 32\\
    hidden dropout & 0.1 & 0.1 & 0.1 & 0.1 & 0.1 & 0.1 & 0.1 & 0.1 \\
    learning rate & 1e-03 & 1e-03 & 1e-03 & 1e-03 & 1e-03 & 1e-03 & 1e-03 & 1e-03\\

    \midrule
    \multicolumn{9}{c}{\textit{\atb}}\\
    \midrule
    batch size & 8 & 16 & 16 & 16 & 8 & 16& 8 & 64\\
    hidden dropout & 0.2 & 0.2 & 0.1 & 0.1 & 0.1 & 0.1 & 0.1 & 0.1 \\
    learning rate & 1e-03 & 1e-03 & 1e-03 & 1e-03 & 1e-03 & 1e-03 & 1e-03 & 1e-03\\
    
    \midrule
    \multicolumn{9}{c}{\textit{\artb}}\\
    \midrule
    batch size & 64 & 64 & 16 & 64 & 32 & 64& 32 & 8\\
    hidden dropout & 0.1 & 0.1 & 0.1 & 0.1 & 0.1 & 0.1 & 0.1 & 0.1 \\
    learning rate & 1e-03 & 1e-03 & 1e-03 & 1e-03 & 1e-03 & 1e-03 & 1e-03 & 1e-03\\
    
    \bottomrule
    \end{tabular}
\end{adjustbox}
    \caption{Best Hyperparameters for Arabic \bert{-based} and T5-based models on all ALUE tasks.}
    \label{tab:hp_cross}
\end{table*}

\begin{figure*}[!htb]
    \centering
    \includegraphics[width=15cm]{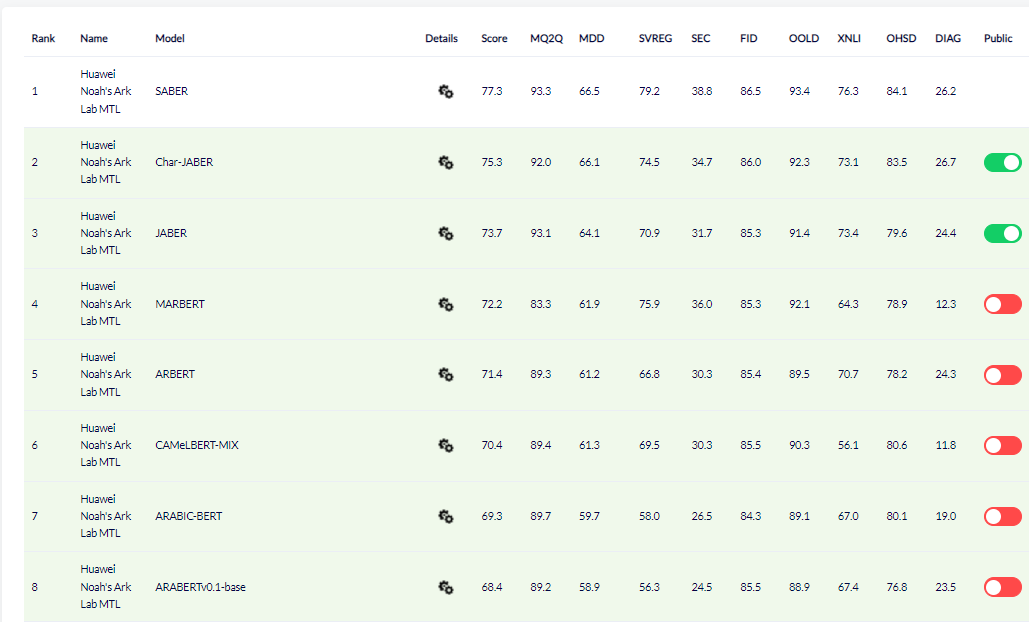}
    \caption{Screenshot of ALUE leaderboard as by 24/05/2022. Red bottoms indicate our private submission baselines which are not visible to the public.}
    \label{fig:leaderboard}
\end{figure*}

\end{document}